\documentclass[letterpaper, 10 pt, conference]{ieeeconf}  % Comment this line out if you need a4paper

\IEEEoverridecommandlockouts                              % This command is only needed if 
\usepackage{amsmath} % assumes amsmath package installed
\usepackage{amssymb}  % assumes amsmath package installed
\usepackage{graphicx}
\usepackage{subcaption}
\usepackage{comment}
\usepackage{cite}
\usepackage{bm}
\usepackage{breqn}
\usepackage{caption}
\usepackage{subcaption}
\usepackage{multirow}
\usepackage[subtle,tracking=normal]{savetrees}

\usepackage{hyperref}
\usepackage[percent]{overpic}
\usepackage{xcolor}
\newcommand{\insertYoutubeLink}{\url{https://youtu.be/elvPv8mq1KM}}  

\title{\LARGE \bf
Learning Human-Robot Handshaking Preferences for Quadruped Robots
}

\author{Alessandra Chappuis, Guillaume Bellegarda, Auke Ijspeert% 
\thanks{
This research is supported by the Swiss National Science Foundation (SNSF) as part of project No.197237. 
The authors are with the BioRobotics Laboratory, Ecole Polytechnique Federale de Lausanne (EPFL).
 {\tt \{firstname.lastname\}@epfl.ch}}
}

\begin{document}
\bstctlcite{MyBSTcontrol}

\maketitle
\thispagestyle{empty}
\pagestyle{empty}

%%%%%%%%%%%%%%%%%%%%%%%%%%%%%%%%%%%%%%%%%%%%%%%%%%%%%%%%%%%%%%%%%%%%%%%%%%%%%%%%
\begin{abstract}
Quadruped robots are showing impressive abilities to navigate the real world. If they are to become more integrated into society, social trust in interactions with humans will become increasingly important. Additionally, robots will need to be adaptable to different humans based on individual preferences. In this work, we study the social interaction task of learning optimal handshakes for quadruped robots based on user preferences. While maintaining balance on three legs, we parameterize handshakes with a Central Pattern Generator consisting of an amplitude, frequency, stiffness, and duration. Through 10 binary choices between handshakes, we learn a belief model to fit individual preferences for 25 different subjects. Our results show that this is an effective strategy, with 76\% of users feeling happy with their identified optimal handshake parameters, and 20\% feeling neutral. Moreover, compared with random and test handshakes, the optimized handshakes have significantly decreased errors in amplitude and frequency, lower Dynamic Time Warping scores, and improved energy efficiency, all of which indicate robot synchronization to the user's preferences. Video results can be found at \insertYoutubeLink.
\end{abstract}

%%%%%%%%%%%%%%%%%%%%%%%%%%%%%%%%%%%%%%%%%%%%%%%%%%%%%%%%%%%%%%%%%%%%%%%%%%%%%%%%
\vspace{-0.12em}
\section{Introduction}
\vspace{-0.06em}

Human handshaking serves as a fundamental tool for social communication and rapport establishment. This behavior requires coordination between the two participants and is influenced by factors such as gender, social context, and interpersonal intimacy \cite{melnyk_physical_2019}. Psychologists have long studied its psychological implications, with research indicating that handshakes can influence perceptions of personality traits, such as warmth and competence~\cite{chaplin2000handshaking}. Moreover, cultural variations in handshake customs underscore the importance of context and social norms in interpreting this gesture~\cite{lafrance1978cultural}. For instance, while firm handshakes are often associated with confidence in Western cultures, other cultures may prioritize different qualities in handshake interactions~\cite{katsumi2017nonverbal}.  
 
\subsubsection{Human-Robot Interaction} In Human-Robot Interaction scenarios, physical touch plays a central role in various applications of social robots as a natural and interactive non-verbal behavior~\cite{breazeal2004designing,lee2006physically}. Robot handshaking has been a topic of interest for showing social and adaptable control and acceptance, towards building trust~\cite{avelino2018power,si2016establish,otterdijk2023shake}. Most existing work on human-robot handshaking uses a full humanoid robot~\cite{prasad2021learning}, or a manipulator arm~\cite{mura2020role}. Handshakes are typically parameterized by several distinct phases including reaching, grasping, and shaking~\cite{prasad_human-robot_2022}. During the shaking phase, there are several important characteristics, including amplitude and frequency of displacements, duration, and firmness and stiffness of the grasp~\cite{knoop2017handshakiness,mura2020role}. Common control strategies to emulate these characteristics with existing robots include Central Pattern Generators (CPGs)~\cite{artem_physical_2013,jouaiti_hebbian_2018} and abstract/neural oscillators~\cite{kasuga2005human,beaudoin_haptic_2019}, imitation learning~\cite{prasad2021learning,falahi2014adaptive}, or deep reinforcement learning~\cite{christen2019guided}.

 \begin{figure}
    \centering
    \includegraphics[width=\linewidth,trim={28cm 22cm 47cm 8cm},clip]{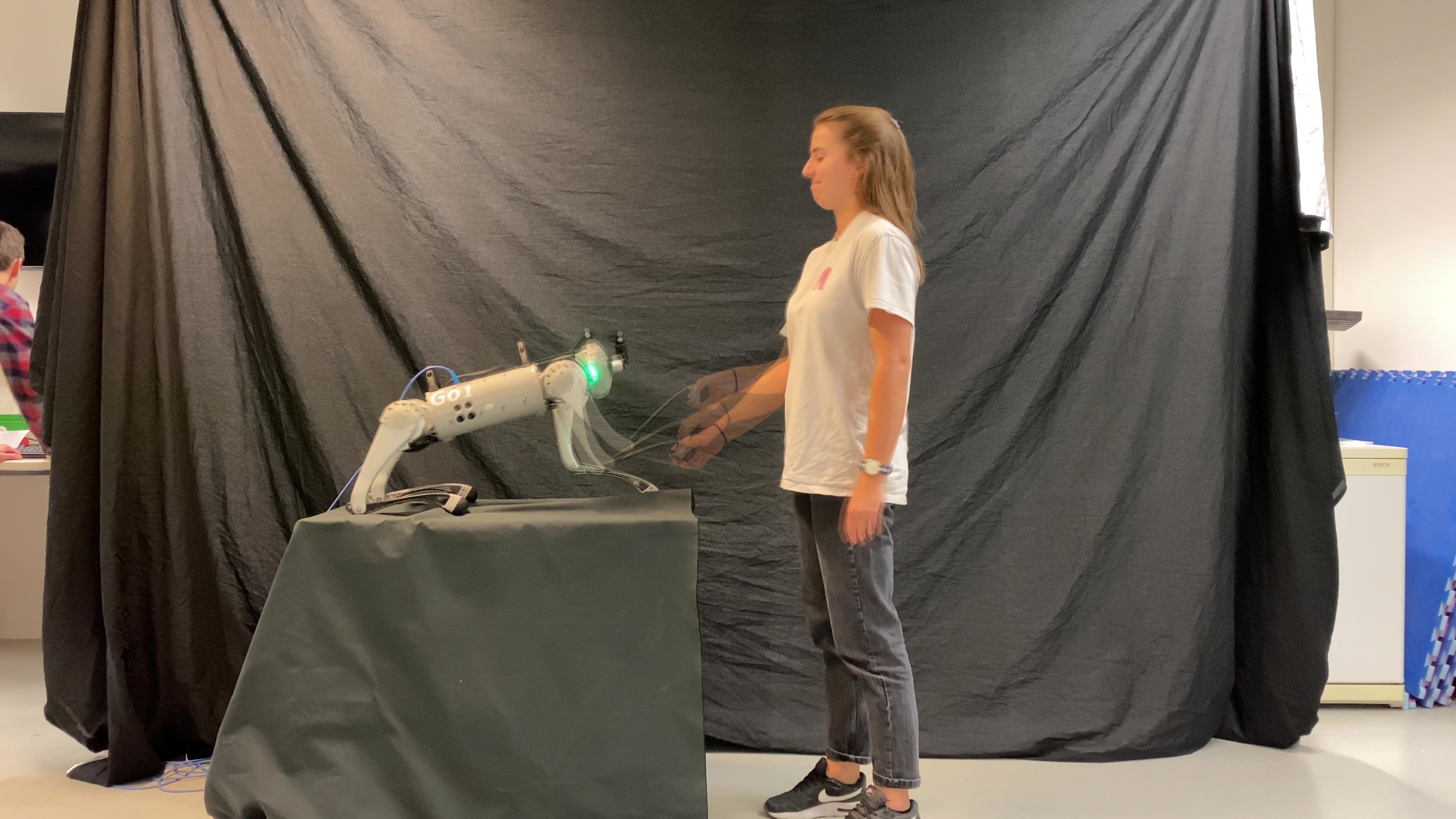}\\
    \vspace{-0.2em}
    \caption{Learning human-preferred handshaking on the Unitree Go1.  %quadruped. 
    }
    \vspace{-1.7em}
    \label{fig:enter-label}
\end{figure}

Most of these works aim at achieving synchronous motion between the human and the robot while minimizing interaction forces \cite{prasad_advances_2020}. For example, Melnyk et al.~\cite{artem_physical_2013} developed a controller based on two coupled oscillators (CPGs) to learn the human rhythm.  Their work focused specifically on learning the frequency of motion within one degree of freedom along the vertical axis. Jouaiti et al.~\cite{jouaiti_hebbian_2018} employed the Hebbian learning mechanism proposed by Righetti et al.~\cite{righetti_dynamic_2006} to synchronize a bio-inspired CPG with an external stimulus.  The CPG was able to accurately learn a new frequency and maintain it once the disturbance stops. Unlike the previous study, they focused on motion in the sagittal plane, resulting in a 2D movement. Beaudoin et al. \cite{beaudoin_haptic_2019} used harmonic oscillator systems with a simple sinusoidal motion to represent movements during handshaking, tracked with an impedance controller. They investigated the impact of various handshake parameters -- including amplitude, frequency, stiffness, and damping -- on the perceived handshake. However, their study did not involve any learning processes.

\subsubsection{Quadruped Robots} In contrast to most systems used for handshaking, quadruped robots are showing impressive abilities to traverse challenging environments~\cite{miki2022learning,bellegarda2022cpgrl}, run at high speeds~\cite{bellegarda2022robust}, jump in difficult terrains~\cite{bellegarda2020robust}, and locomote over dynamic parkour obstacles~\cite{shafiee2023deeptransition,shafiee2023puppeteer,cheng2023parkour}. However, there are still limited examples of social interactions between quadruped robots and humans, which will become increasingly important as quadrupeds become more integrated into society. 

\subsubsection{Contribution} In order to study handshaking with common quadruped robots such as the Unitree Go1~\cite{unitreeGO1}, we note two important considerations: 1) they must lift one of their feet to handshake while maintaining balance to avoid falling, and 2) they have point feet rather than hands or grippers, which is atypical for handshaking tasks. We solve the first consideration with inspiration from dogs, who normally sit on their rear legs while ``handshaking'' (i.e.~giving a paw), and maintain balance with the rear legs and their remaining front foot. For the second consideration, we parameterize a handshake with the following common parameters: amplitude displacement, frequency, stiffness, and a fixed duration. 

Humans exhibit a wide range of different handshaking styles depending on many factors, including their culture, gender, or social contexts, for example using different handshaking styles when greeting vs.~consoling~\cite{melnyk_physical_2019,tagne_physical_2016}. Therefore, in contrast to other works studying human-robot handshaking, we aim to develop an adaptable system that can fit individual user preferences, ideally rapidly within few interactions with the robot. In this context, preference learning~\cite{biyik_active_2020,akrour2012april} has been used to learn user preferences for varying tasks including preferences in driving styles~\cite{sadigh_active_2017,basu2017you} and human exoskeleton gait parameters~\cite{tucker2020preference,ingraham2022role,lee2023user}. In preference learning, the user is exposed to two different experiences at a time, and makes a binary choice between which one they prefer. This continues in a loop for a number of repetitions, typically until the underlying belief converges to the inferred optimal parameters for that user. 

In summary, in this paper we present a framework for learning individual human-robot handshaking preferences for quadruped robots. The robot first sits on its hind legs and extends its front right foot to prepare for a handshake. Subsequently, when a human grasps the foot, it begins a handshake parameterized by an amplitude, frequency, stiffness, and fixed duration. By iteratively showing the human two handshakes with different parameters and having them select their favorite between the two, we learn a belief of the human's preferred parameters with preference learning. After 10 trials, the human is shown their identified optimized belief, which we then validate with ablation studies by perturbing individual parameters. Our results from learning preferences for 25 subjects show that this is an effective strategy, with 76\% feeling happy with their identified optimal handshake parameters, and 20\% feeling neutral. Moreover, compared with random and test time handshakes, the optimized handshakes have significantly decreased errors in intended amplitude and frequency, lower Dynamic Time Warping scores, and improved energy efficiency. 

The rest of this paper is organized as follows. In Section~\ref{sec:method} we present our methods for generating and parameterizing quadruped handshakes, as well as preference learning for optimizing these parameters for each individual. In Section~\ref{sec:exp_metrics} we detail our experimental procedure for learning user-preferred handshaking parameters. Section~\ref{sec:results} discusses results and analyzes metrics from learning handshaking parameters for 25 subjects, and a brief conclusion is given in Section~\ref{sec:conclusion}.

%%%%%%%%%%%%%%%%%%%%%%%%%%%%%%%%%%%%%%%%%%%%%%%%%%%%%%%%%%%%%%%%%55
\section{Methods}
\label{sec:method}
\begin{figure*}
    \centering
    \includegraphics[width=0.95\linewidth]{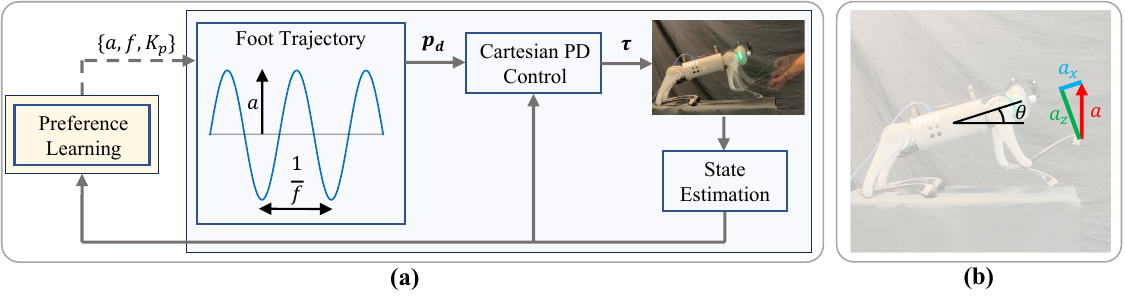}\\
    \vspace{-0.8em}
    \caption{Control diagram for mapping preferred parameters to joint torques for handshakes with a quadruped robot. \textbf{(a)}: Optimized handshake parameters amplitude ($a$), frequency ($f$), and Cartesian stiffness ($K_p$) create a desired foot trajectory $\bm{p}_d$ which is tracked with Cartesian PD control. The solid lines operate at 1 kHz, while the dotted line indicates the new handshake parameters, which are sent when a grasp is detected. \textbf{(b)}: the world frame amplitude $a$ is mapped to the robot frame based on the robot pitch $\theta$. 
    }
    \vspace{-0.5em}
    \label{fig:ctrl}
\end{figure*}

In this section we describe our framework for generating and parameterizing handshakes, configuring and mapping these parameters to be executed with a quadruped robot, and the preference learning procedure for identifying optimal user handshake parameters. A high-level control diagram is shown in Figure~\ref{fig:ctrl}, and we explain all components below.

\begin{figure*}
    \centering
    \includegraphics[width=0.97\linewidth]{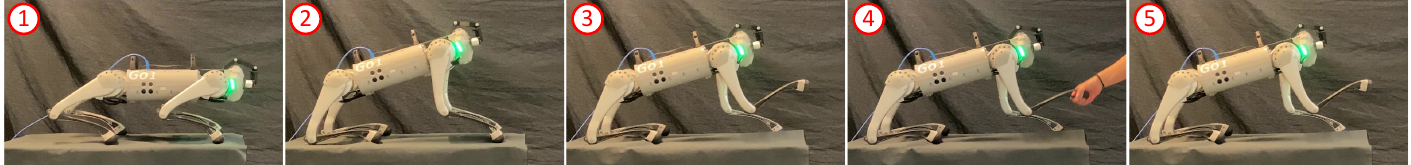} \\
    \vspace{-0.2em}
    \caption{Robot positions during handshaking experiments. \textbf{(1)}: standing at rest.
    \textbf{(2)}: sitting on rear legs. \textbf{(3)}: raising the front right foot and waiting for a grasp. \textbf{(4)}: performing a user handshake. \textbf{(5)}: returning to the nominal shaking position and waiting for the next grasp.
    }
    \label{fig:robot_snapshots}
    \vspace{-1.5em}
\end{figure*}

\subsection{Generating and Parameterizing Handshakes}
There are several parameters that are important for characterizing a handshake. These include frequency, amplitude, stiffness, and duration. Due to the oscillatory nature of handshakes and their parameters, abstract oscillators  provide an intuitive formulation for representing and generating a handshake. A number of abstract oscillators have previously been used for handshakes \cite{kasuga2005human,artem_physical_2013,jouaiti_hebbian_2018,beaudoin_haptic_2019}, and to generate rhythmic locomotion skills for quadruped robots~\cite{righetti08,bellegarda2022cpgrl,bellegarda2024visual,bellegarda2024quadruped}. Similarly, here we consider the following parameterization for vertical oscillations to define a handshaking trajectory with: 
\begin{equation}
\label{eq:handshake}
\begin{split}
    z(t) &= z_{nom} + a \cdot \sin(2\pi f \cdot t)
\end{split}
\end{equation}
where $a$ is the amplitude in \texttt{cm},  $f$ is the frequency in Hz, $z_{nom}$ is the nominal $z$ height around which the oscillation takes place, and $t$ is time. Since the robot will be ``sitting'' on its rear limbs during the handshake, we ensure the handshake occurs in the world $z$ axis and robot $XZ$ plane by mapping the amplitude into $x$ and $z$ components with: 
\begin{align}
    a_x &= -a \cdot \sin(\theta) \\
    a_z &= a \cdot \cos(\theta)
\end{align} 
where $\theta$ is the pitch of the body, as visualized in Figure~\ref{fig:ctrl}-b. The desired foot trajectory $\bm{p}_d (t)$ in the 
robot leg frame coordinates is thus:
\begin{align}
    \bm{p}_d (t) = \begin{bmatrix}
        x_{nom} + a_x \cdot \sin(2\pi f \cdot t) \\
        y_{nom} \\
        z_{nom} + a_z \cdot \sin(2\pi f \cdot t)
    \end{bmatrix}
\end{align}
where $(x_{nom}, y_{nom}, z_{nom})$ are the nominal foot position coordinates in the leg frame at the beginning of a handshake.

%%%%%%%%%%%%%%%%%%%%%%%%%%%%%%%%%%%%%%%%%%%%%%%%%%%%%%%%%%%%%%%%%55
\subsubsection{Control}  \label{sec:cartesianPD}
Figure~\ref{fig:ctrl}-a shows an overview of the control diagram. The desired foot position is mapped to joint commands with a Cartesian PD controller. 
The foot position is thus mapped to torques and tracked at the joint level with the following controller for the front right leg: 
\begin{align}
    \bm{\tau} &= \bm{J}(\bm{q})^\top \Bigl[ \bm{K}_{p} \left(\bm{p}_{d} - \bm{p} \right) - \bm{K}_{d} \left( \bm{v} \right)  \Bigr] - \bm{K}_{d,joint} (\bm{\dot{q}})
    \label{eqn:leg_ff}
\end{align}
where $\bm{J}(\bm{q})$ is the foot Jacobian at joint configuration $\bm{q}$, $\bm{K}_p$ and $\bm{K}_d$ are diagonal matrices of proportional and derivative gains in Cartesian coordinates to track the desired foot positions $(\bm{p}_d)$ with zero desired foot velocity $(\bm{v})$ in the leg frame. We add a small joint damping term for stability in the hardware experiments. The stiffness $\bm{K}_{p}=k\bm{I}_3$ is a user-parameter we will learn, and we set $\bm{K}_{d}=0.02\bm{K}_{p}, \  \bm{K}_{d,joint}=0.8\bm{I}_3$ similarly to our previous works~\cite{bellegarda2022cpgrl,bellegarda2024visual,bellegarda2024quadruped}, where $\bm{I}_3$ is the $3 \times 3$ identity matrix.

%%%%%%%%%%%%%%%%%%%%%%%%%%%%%%%%%%%%%%%%%%%%%%%%%%%%%%%%%%%%%%%%%55
\subsection{Handshake Overview}
Performing a handshake with a quadruped robot has several necessary steps. In particular, the quadruped needs to raise one of its feet while maintaining balance, i.e. keeping its center of mass within the support polygon created with its remaining limbs. The quadruped must also detect when a handshake should begin, either visually or by sensing a human has grasped its foot. An overview of the steps and components is illustrated in Figure~\ref{fig:robot_snapshots}; 
namely, the quadruped:
\begin{enumerate} 
    \item Begins standing at rest
    \item Sits on its rear legs
    \item Lifts its front right foot to a nominal shaking position, and waits until it detects a foot grasp (waiting state) \label{wait}
    \item Performs the handshake for 3 seconds (shaking state) \label{handshake}
    \item Returns to the nominal shaking position \label{return}
\end{enumerate}
Steps \ref{wait}-\ref{return} are repeated during the iterative process for each handshake, switching between the waiting and shaking state. 

Detecting when a user grasps the leg is essential to correctly initiate the handshake. As the foot is grasped and thus perturbed from its nominal position, there is an increase in the joint torques as the controller tries to regulate the foot back to its nominal position. We define a threshold of 150\% of the nominal joint torques in at least two out of three of the joints in the front right leg to determine that a user has grasped the leg, and thus initiate the handshaking. 

The Cartesian PD controller explained in Section~\ref{sec:cartesianPD} is employed to track the handshake trajectory. This choice enables the robot leg to vary the compliance with the user motions, meaning it can be more or less shaken based on the preferences (i.e.~of $\bm{K}_p$).

%%%%%%%%%%%%%%%%%%%%%%%%%%%%%%%%%%%%%%%%%%%%%%%%%%%%%%%%%%%%%%%%%55
\subsection{Active Preference-Based Reward Learning}
The primary objective of reward function learning is to learn a motion from user preferences with minimal data \cite{sadigh_active_2017}. The user gives relative preferences among multiple trajectories,  rather than relying on predefined reward structures, to facilitate the learning of the reward function. Here we use APReL -- a library for active preference-based reward learning~\cite{biyik_aprel_2022}.

\begin{figure*}[!t]
    \centering
    \vspace{-1.3em}
    \includegraphics[width=0.95\linewidth]{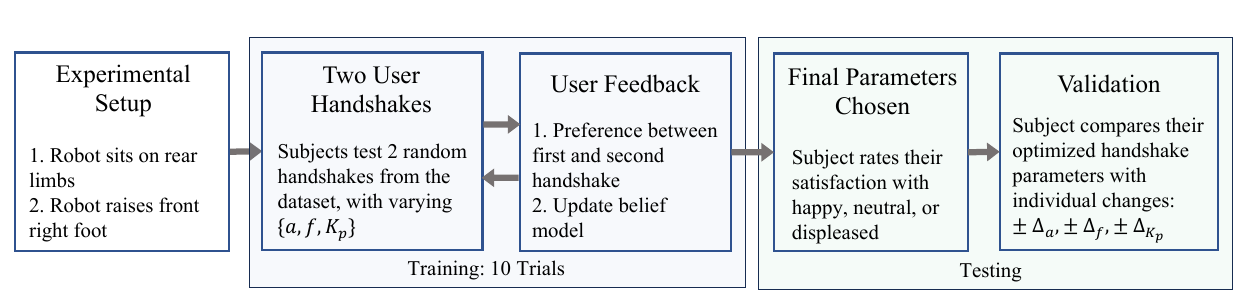}\\
    \vspace{-0.65em}
    \caption{\textbf{Experimental procedure.} The robot first sits on its rear legs and lifts its front right foot to prepare for user handshakes. The training block (shaded blue) consists of 10 trials of two sets of random handshake parameters each, with the user giving feedback on which handshake they prefer. The belief model is updated after each pair of handshakes to reflect the user preferences. After 10 trials, the user is shown the parameters consisting of their identified preferences, and rate their satisfaction with happy, neutral, or displeased. Lastly, we perform a validation study where the user alternates between their optimally identified parameters, and high/low perturbations to a single parameter at a time. }
    \label{fig:exp_setup}
    \vspace{-1.5em}
\end{figure*}

As an overview, we systematically present two different trajectories (i.e.~handshakes) to the user, and ask their preference between the two. Through iterative presentation of various queries (i.e.~trajectory parameter sets), the algorithm learns the reward function associated to the user, and identifies the trajectory that maximizes this reward. The optimized trajectory should then correspond to the user's preferred handshake parameters. Here we consider three key features to be learned which describe a handshake: the amplitude and frequency of the oscillations, and the stiffness of the controller.
The reward function that we aim to learn is expressed as a linear combination of trajectory parameters:
\begin{equation}
    R(\xi)=\pmb{\omega}^{\top} \Phi(\xi) .
    \label{eq:reward}
\end{equation}
In this equation, $R(\xi)$ is the reward associated with a trajectory, $\pmb{\omega}$ denotes the weights, and $\Phi(\xi)$ represents the trajectory features. As the robot trajectory features (handshake parameters) $\Phi(\xi)$ are known, the task consists of estimating the weights $\pmb{\omega}$ based on the user's preferences, and to optimize the handshake parameters accordingly. The following modules are used to achieve the learning process: 

\noindent \textbf{Initial trajectory set generation.} Random handshake trajectories are generated from within the identified handshake parameter ranges. The trajectories are later proposed to the user, serving as input for the reward function learning process.

\noindent \textbf{Comparison queries and user feedback.} We employ comparison-type queries, where each query contains two trajectories randomly selected from the initial trajectory set. The user is then asked to provide binary feedback on which handshake they preferred. 

\noindent \textbf{User response model.}  The learning process relies on a user response model that provides probabilities of trajectory selection based on the current query and the reward function. In this study, the soft-max model is used: 
\begin{align}
    P(c = \xi_c)  = \frac{\exp(R(\xi_c))}{\sum_{j}\exp(R(\xi_j))}
\end{align}
where $P(c = \xi_c)$ is the probability of the user choosing trajectory $\xi_c$, and $R(\xi_j)$ is the reward function associated to the trajectory $\xi_j$. 

\noindent \textbf{Belief distribution and update.} Bayesian learning is performed through a belief distribution module, incorporating user feedback and the user response model. A sampling-based posterior distribution model, using the Metropolis-Hastings algorithm (a Markov chain Monte Carlo method), iteratively updates the weights of the reward function based on the user preferences. The initial beliefs are sampled from a normal distribution. Then, the algorithm obtains a new user choice, and generates a candidate point from the current user distribution model. If the candidate point satisfies the Metropolis-Hastings criteria, it is accepted and the user distribution model is updated. For additional details, we refer the reader to~\cite{sadigh_active_2017,biyik_aprel_2022}. 

%%%%%%%%%%%%%%%%%%%%%%%%%%%%%%%%%%%%%%%%%%%%%%%%%%%%%%%%%%%%%%%%%55
\section{Experimental Procedure and \\ Evaluation Metrics}
\label{sec:exp_metrics}

\subsection{Experimental Procedure}
To evaluate the performance of the previously explained methods, we conducted experiments with 25 participants (11 female and 14 male) aged 20-35 years old. The robot was positioned on an elevated box to ensure that the handshake occurred at a natural height for human participants. Each handshake lasted 3 seconds, which we determined to be an appropriate duration for participants to distinguish the differences between two handshakes. 

Table~\ref{tab:range} shows the ranges of the possible handshake parameters of amplitude, frequency, and stiffness. For each user, we generated a set of 12 random handshakes uniformly distributed within these ranges, as well as 3 passive handshakes of variable stiffness. We selected a total of 15 distinct handshakes to ensure an effective balance between variety and distinguishability. An excessive number of handshakes could result in subtle differences between handshakes that make it difficult for users to discern and express clear preferences. Conversely, too few parameter sets might result in an insufficient range of options, which could compromise the algorithm's ability to learn the user's preferences. For the passive handshakes, the amplitude and frequency are set to zero, while the Cartesian $K_p$ gain is assigned values of 30, 115, and 200, respectively. These passive handshakes of variable stiffness are chosen to investigate whether humans generally prefer active or passive handshakes, and with low or high stiffness, similar to~\cite{mura2020role}. 

\begin{table}
\centering
\caption{Handshake parameter ranges.}
\vspace{-0.4em}
\begin{tabular}{ c | c | c }
 \hline
 Parameter & Min value & Max value \\
 \hline
 Amplitude ($a$) [cm]     & 1  & 10   \\
 Frequency ($f$) [Hz]     & 1  & 3.5  \\
 $K_p$ gain ($k$)         & 30 & 200  \\
 \hline
\end{tabular}
\label{tab:range}
\vspace{-1.8em}
\end{table}

An overview of the experimental procedure is shown in Figure~\ref{fig:exp_setup}. For each subject, the experiment consisted of three phases:

\begin{enumerate}
    \item \textbf{Preference learning:} Participants engaged in 10 handshake pairwise comparisons to enable the preference learning algorithm to learn their preferences, where the handshakes were randomly selected. 
    \item \textbf{Optimized handshake experience:} Participants experienced and rated the optimized handshake generated based on their learned preferences.
    \item \textbf{Validation study:} To assess the quality of the estimated parameters, we conducted an ablation study of 6 handshake comparisons in which users were presented with two handshake options and asked their preference. One option was their optimized handshake, while the other was a modified version with one of the three parameters altered as follows: amplitude $\pm$ 2 cm, frequency $\pm$ 0.5 Hz, or $K_p$ $\pm$ 40.
\end{enumerate}

By following this experimental procedure, we aim to gain insights into the effectiveness of preference learning in personalizing handshakes, and the impact of each parameter on the overall handshake experience.

\subsection{Evaluation Metrics} 
To assess the performance of our preference learning approach, we employ several evaluation metrics, focusing on the overall user satisfaction, mismatches between parameters and human feedback, and global similarity between the performed and desired handshakes.
\begin{itemize}
    \item \textit{Overall User Satisfaction}: The first and most important metric is user satisfaction. After the user is shown the 10 pairwise handshake comparisons, the identified optimized handshake parameters are presented to the user. The user then qualitatively rates their experience with their identified handshake parameters among: `happy', `neutral', and `displeased'. 
    \item \textit{Amplitude and Frequency Error}: We calculate the relative error in amplitude and frequency between the learning phase handshakes, slightly modified handshakes from the ablation study, and optimized handshakes. To compute the relative error, we compare the desired average amplitude and frequency with the actual trajectory performed in conjunction with the user for each handshake type.
    \item \textit{Mean (Absolute) Torque}: We compute the mean total absolute torque during each 3 second handshake by summing the absolute mean torque of each of the three motors in the front right leg. A larger mean torque can correspond either to high amplitude/frequency parameters, or to a lack of synchronization between the human and robot, as the human significantly deviates the foot from its intended trajectory causing large position errors, which the robot attempts to regulate back to the nominal trajectory with larger torques.  
    \item \textit{Dynamic Time Warping (DTW)}: To assess the global similarity between the performed and desired handshakes, we utilize Dynamic Time Warping (DTW). DTW is an algorithm for measuring the similarity between two temporal sequences which may vary in speed or length \cite{berndt1994using}. The higher the DTW score, the greater the distance between the two signals, indicating a lower level of similarity between the intended robot trajectory and the human in-the-loop handshake trajectory.
    \item \textit{Phase Locking Value (PLV)}: The PLV is associated with the absolute phase difference between the robot and human handshake trajectories. Values closer to 0 indicate a lack of synchrony between the two signals. Conversely, a constant phase difference, associated with similar motion between the trajectories, results in a PLV close to 1 \cite{melnyk_physical_2019}.
\end{itemize}

%%%%%%%%%%%%%%%%%%%%%%%%%%%%%%%%%%%%%%%%%%%%%%%%%%%%%%%%%%%5
\section{Results and Discussion}
\label{sec:results}
In this section we present results from using preference learning to optimize 25 individual users' handshaking preferences with a quadruped robot. We are specifically interested in the following questions: 
\begin{itemize}
    \item What is the satisfaction rate among users regarding their identified preferred handshaking parameters? 
    \item Are there differences between male and female participants in final parameter preferences? 
    \item Do the identified optimal parameters improve metrics in performing handshakes in terms of  amplitude/frequency error, DTW, and mean torque? 
\end{itemize}

\begin{figure}
    \centering
    \vspace{-0.3em}
    \includegraphics[width=0.95\linewidth]{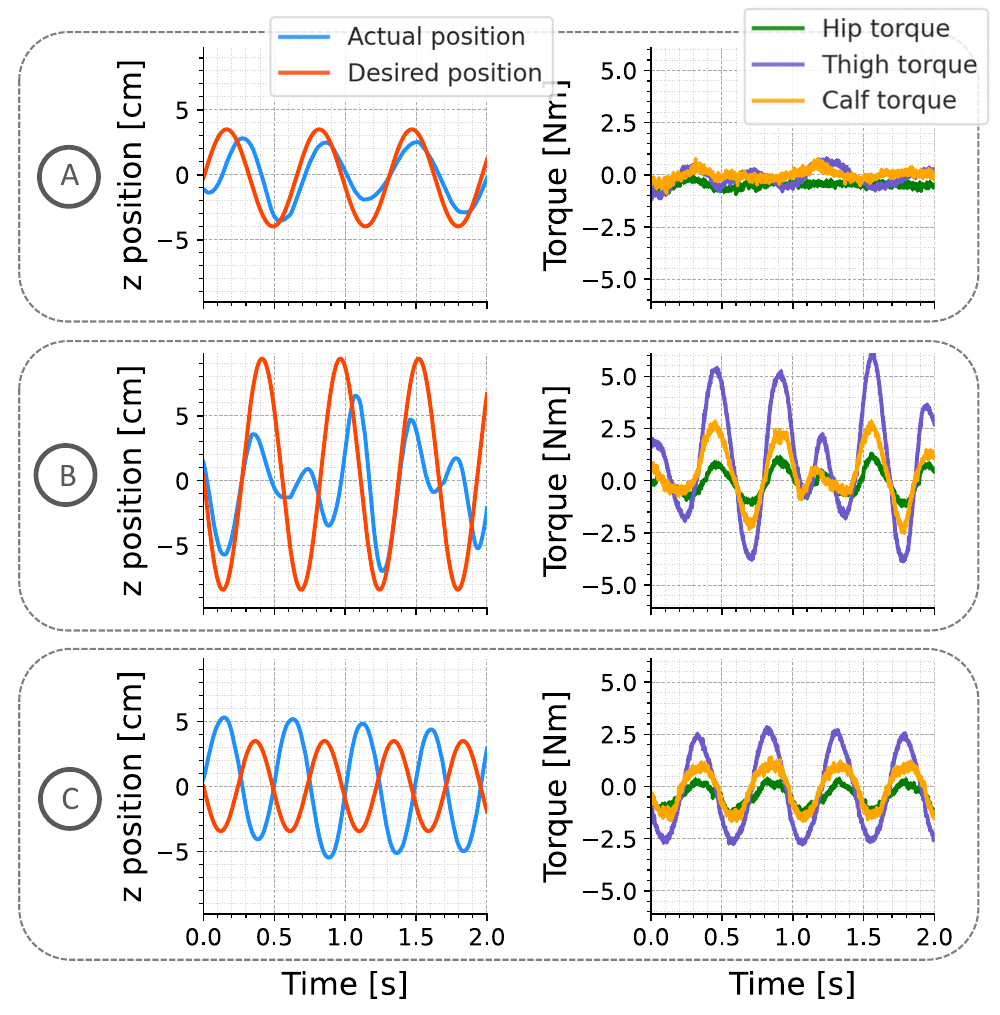}\\
    \vspace{-0.5em}
    \caption{\textbf{Comparison of three distinct handshakes.} Plots of the foot position along the z-axis (left) and the joint torques (right) as a function of time. For each handshake the corresponding parameters are indicated: $\{a, f, K_p\}$. \textbf{(A)}: synchronous motion between the user and the robot, $\{3.5, 1.53, 114.87\}$. \textbf{(B)}:~asynchronous motion, $\{9.4, 1.81, 143.71\}$. \textbf{(C)}: out-of-phase motion, with the robot and the user moving in opposite intended directions, $\{3.5, 2.05, 73.3\}$.
    }
    \vspace{-1.5em}
    \label{fig:threeHs}
\end{figure}

%%%%%%%%%%%%%%%%%%%%%%%%%%%%%%%%%%%%%%%%%%%%%%%%%%%%%%%%%%%%%%%%
\subsection{Sample Handshakes and Synchrony}
We begin our discussion by first observing three different sample handshake parameterizations, and the actual foot trajectory during the handshake, shown in Figure~\ref{fig:threeHs}. Notably, varying synchronization can be observed, where the user and robot are almost fully in synchrony for handshake A, while fully out of phase for handshake C. Handshake B shows the robot and user greatly differ in both amplitude and frequency, which combined with a large $K_p$ leads to high torques as the robot fights against the human's shaking. In comparison, a synchronized handshake (A) results in very low torque, as the robot and human move together in synchrony while minimally deviating the intended robot trajectory. By observing the amplitude and frequency errors for the three handshakes, we observe a mean amplitude error of 33\% for handshake A, 51\% for handshake B, and 47\% for handshake C. Additionally, the frequency errors are 3\%, 42\% and 3\% respectively. It is important to note that considering only the frequency error might lead to the incorrect assumption that both handshakes A and C represent synchronous motions. However, when taking into account the DTW scores, which are 12.62m, 51.71m, and 31.07m respectively, it becomes apparent that the score for handshake C cannot be associated with a synchronous motion when compared to handshake A. This highlights the importance of combining multiple metrics, such as amplitude and frequency errors, as well as DTW scores, to accurately evaluate the similarities between the robot and human trajectories during handshake interactions. 

\begin{figure}[t]
    \centering
    \includegraphics[width=0.93\linewidth,trim={0.1cm 0 0 0.1cm},clip]{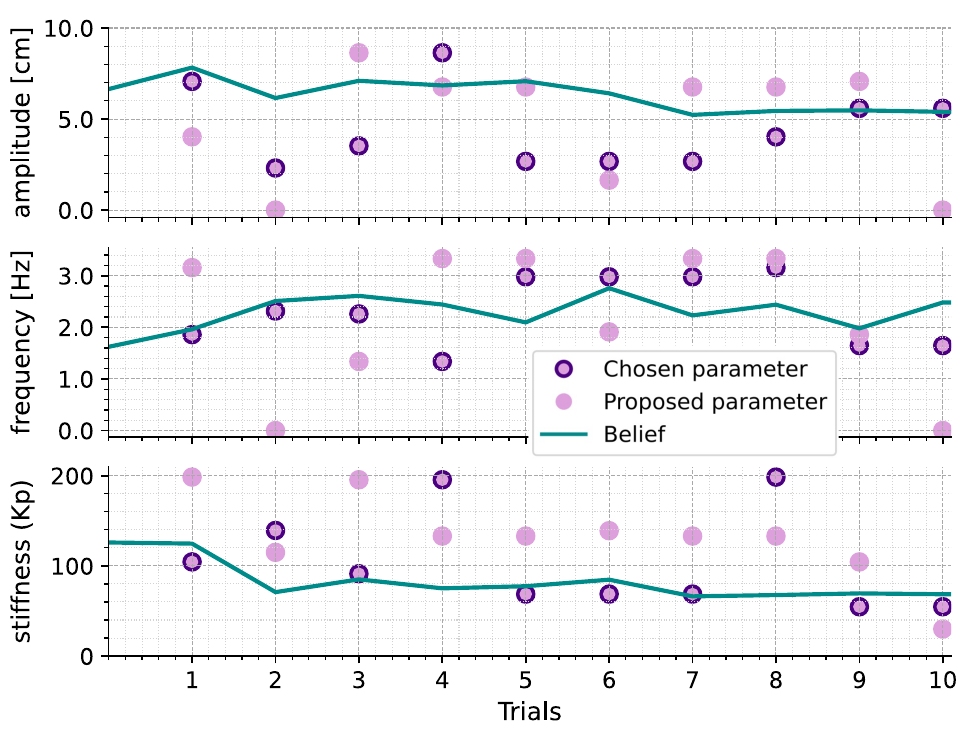}\\
    \vspace{-0.85em}
    \caption{\textbf{Preference learning for one experiment.}  The proposed amplitude, frequency, and stiffness parameters, user choices, and belief evolutions are shown across 10 learning trials. 
    } 
    \label{fig:training1}
    \vspace{-2em}
\end{figure}

%%%%%%%%%%%%%%%%%%%%%%%%%%%%%%%%%%%%%%%%%%%%%%%%%%%%%%%%%
\subsection{Learning Process and User Satisfaction}
\begin{figure}[!t]
    \centering
    \includegraphics[width=0.93\linewidth,trim={0.1cm 0 0 0.1cm},clip]{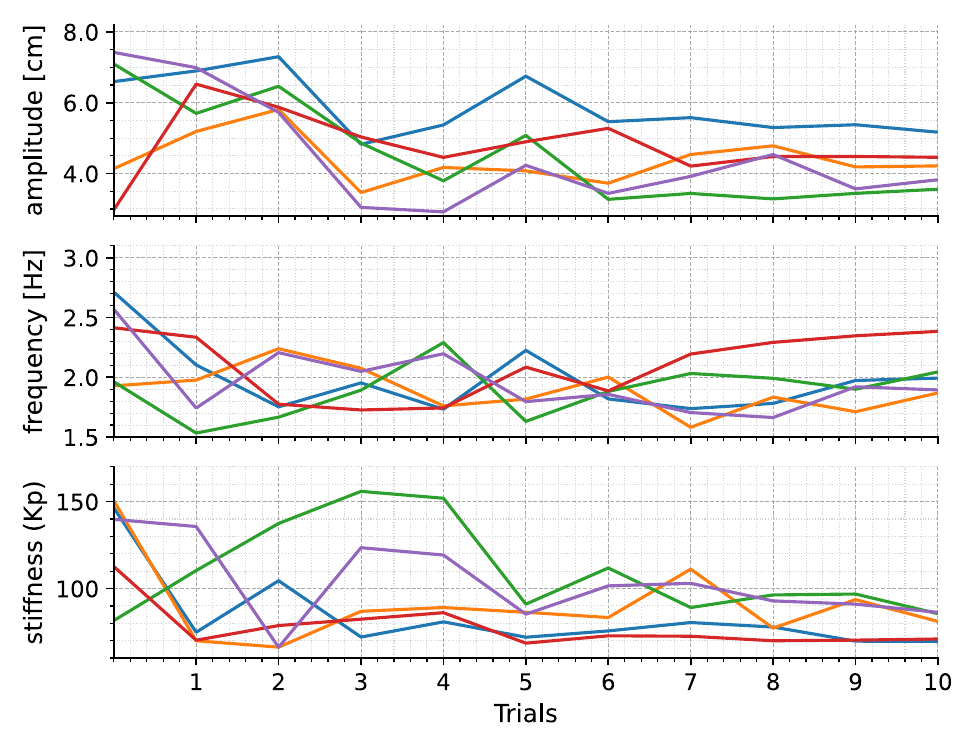}\\
    \vspace{-0.85em}
    \caption{
    Evolution of beliefs for amplitude, frequency, and stiffness for 5 subjects.
    }
    \label{fig:all_beliefs}
    \vspace{-0.6em}
\end{figure}

Figure \ref{fig:training1} shows the learning process for a single user with the randomly proposed handshake parameters, the subject's choices, and the internal belief evolution over the trials. It visualizes how the preference learning algorithm adapts to the user feedback and adjusts its estimate of the optimal handshake parameters throughout the interactions. The initial belief values are randomly sampled from a normal distribution, which are shown as trial 0. As the user selects trajectory preferences during the trials, the algorithm updates the belief values accordingly. Over time, the algorithm converges to a final set of parameters which reflect the user preferences. Throughout the learning process, the algorithm demonstrates the ability to adjust the estimated optimal handshake parameters based on the user's selections. Figure \ref{fig:all_beliefs} shows the evolution of the beliefs for each handshake parameter across the learning processes for 5 different user experiments. The plot shows distinct trajectories for each belief as they converge towards individual subject preferences. 

Table \ref{tab:final_param} presents the mean and standard deviation values for each of the 25 subjects' individually-preferred parameters. Interestingly, the mean beliefs for female participants indicate larger amplitude and stiffness preferences compared to male participants, while males and females both converged to the same mean frequency. 

\subsubsection{Overall Satisfaction} For user satisfaction, 19/25 (76\%) of the participants expressed happiness with the personalized handshake. Additionally, 5/25 (20\%) of the participants viewed their identified handshake parameters as neutral, while only 1/25 (4\%) participants were displeased with the resulting handshake. These results are consistent with those of the validation study: across all ablation tests, the subjects preferred their optimally identified handshake parameters in 71.5\% of the trials, compared with the slightly modified ones. These numbers indicate a high level of user satisfaction with the personalized handshake generated by the preference learning algorithm, after only 10 learning trials. The high level of satisfaction suggests that the algorithm can improve human-robot interactions by providing more natural and enjoyable handshaking experiences.

%%%%%%%%%%%%%%%%%%%%%%%%%%%%%%%%%%%%%%%%%%%%%%%%%%%%%%%%%
\subsection{Evaluation Metrics}
Table~\ref{tab:metrics} presents the metrics used to evaluate three distinct categories of handshakes: \textbf{learning phase} handshakes, which correspond to the 10 learning trials with the user performing 20 total randomly generated handshakes; \textbf{testing} handshakes, consisting of the six handshakes during the validation study with individual parameter changes; and the \textbf{optimized} handshakes, generated using the learned parameters. Each metric is divided into female, male, and total categories. 

\begin{table}
\vspace{-0.1em}
\centering
\caption{Mean and standard deviation of optimally identified handshake parameters after learning the preferences for 25 subjects.}
\vspace{-0.35em}
\resizebox{1\linewidth}{!}{
\begin{tabular}{  c | c | c | c }
     \hline
     Parameter & Female & Male & Total\\
     \hline
     Amplitude [cm]  & 5.21 $\pm$ 1.09 & 4.56 $\pm$ 0.95 & 4.94 $\pm$ 1.15 \\
     Frequency [Hz]  & 2.10 $\pm$ 0.47 & 2.12 $\pm$ 0.34 & 2.11 $\pm$ 0.39  \\
     Stiffness       & 93.16 $\pm$ 23.32 & 83.82 $\pm$ 16.27 & 91.02 $\pm$ 21.72\\
     \hline
\end{tabular}
}
\vspace{-2.5em}
\label{tab:final_param}
\end{table}

\subsubsection{Overall Metrics} To assess similarities and errors between the desired handshake trajectories and the actual foot trajectories performed due to the interaction with the users, we analyze the amplitude and frequency error, as well as the DTW score. When examining the total participant group (T) in Table~\ref{tab:metrics}, we observe a 15\%  decrease in amplitude error between the learning phase and testing handshakes, followed by a further 7\% reduction in amplitude error between the testing and optimized handshakes. The frequency error also shows a similar trend, with a total decrease of 25\% from learning phase to optimized handshakes. However, it is worth noting that the frequency error did not significantly decrease during the testing handshakes. The optimized handshakes were performed with a residual error of 49\%  in amplitude and 14\% in frequency. The residual error in amplitude could indicate that users prefer a different amplitude than the one performed by the robot, resulting in higher interaction forces that make the handshake feel more engaging (assuming the frequency error is minimal). These higher interaction forces from the amplitude error can also be interpreted as another way of changing the stiffness. In other words, a perfect synchronization in both amplitude and frequency might not provide the desired interaction force that users expect from a handshake. In human-human handshaking, this interaction force plays a crucial role, even conveying information such as gender and personality traits~\cite{orefice_lets_2016}. 

The DTW score also follows the same trend for all users (T) with a decrease from the learning phase to the optimized handshakes. This further supports the previous findings on decreasing the amplitude and frequency errors. Regarding the mean torque, the values reported in Table~\ref{tab:metrics} correspond to 20 of the 25 participants. The results of 5 participants were excluded from the analysis, as their mean torque drastically increased during the optimized handshakes. This can be attributed to several factors, with the most significant one being that these users progressively shifted their shaking position, ending up far away from the robot's nominal position during the optimized handshakes. This shift in position, combined with other factors such as poor synchrony and high preferred parameter values, contributed to the increase in the torque values. This highlights the importance of maintaining a consistent handshake position throughout the whole experiment. For the other participants, a decrease of 17\% is observed between the learning phase and optimized handshakes, further demonstrating a high learned synchrony between the user and the robot.

\begin{figure}[t]
    \centering
    \vspace{-0.3em}
    \includegraphics[width=0.95\linewidth,trim={0.2cm 0.5cm 1cm 1cm},clip]{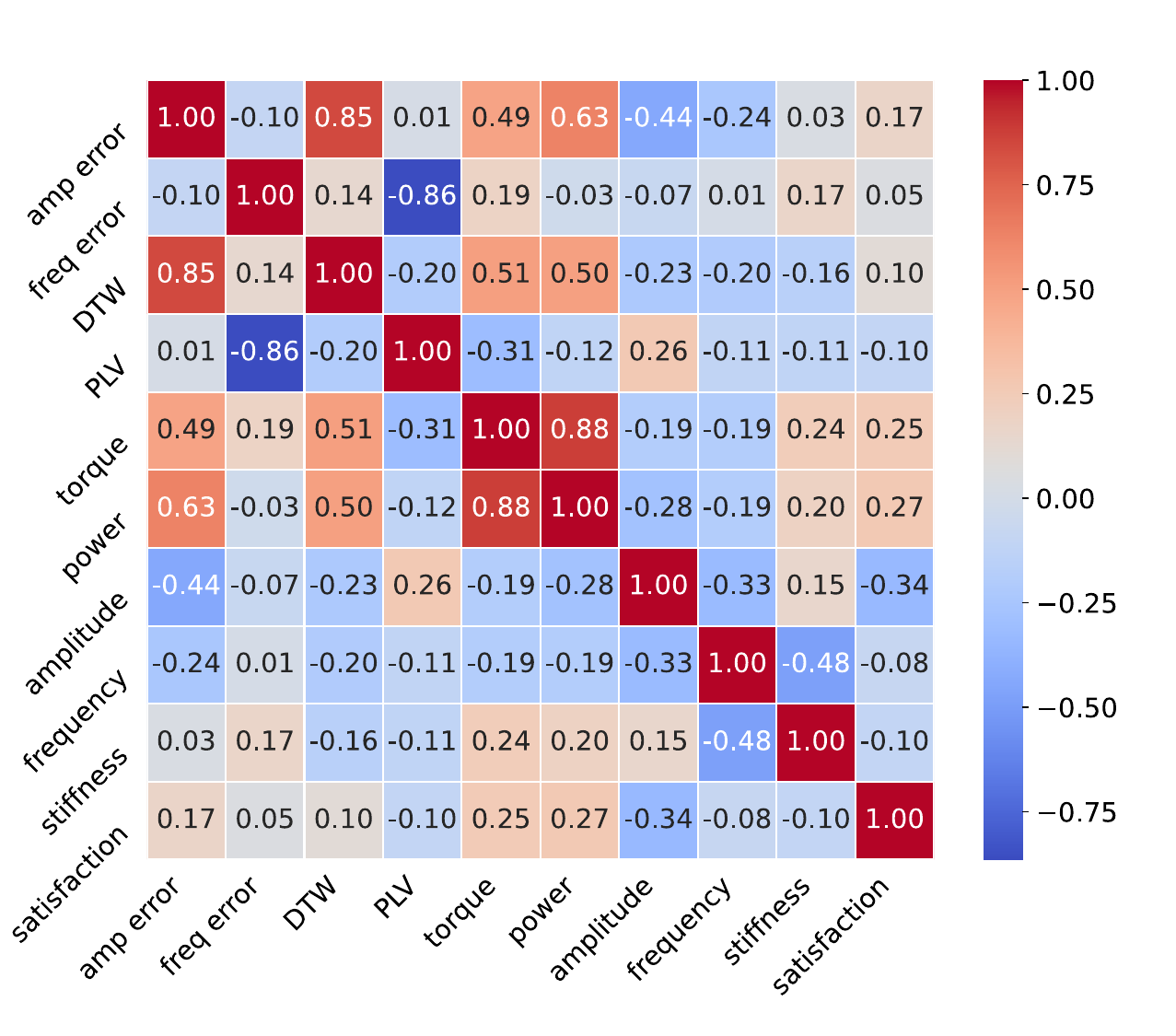}\\
    \vspace{-0.45em}
    \caption{Correlation matrix between the metrics and the optimized handshake parameters, computed using Pearson's correlation coefficient. 
    }
    \label{fig:corr_matric}
    \vspace{-2em}
\end{figure}

\subsubsection{Gender Differences} When comparing gender differences in Table~\ref{tab:metrics}, we can observe that female participants initially have larger amplitude and frequency errors, as well as higher DTW scores, compared to male participants. However, these differences significantly decrease when comparing the optimized handshakes: female participants exhibited lower amplitude errors and DTW scores than male participants in optimized handshakes, while showing similar frequency errors. An interesting result emerges when analyzing the mean torque in the three handshake categories. Male participants consistently showed lower average torque values compared to the female participants, with differences ranging from 20\% (learning phase) to 15\% (optimized), implying better synchrony with the robot. 

\subsubsection{Metric Correlations} Figure \ref{fig:corr_matric} presents the correlation scores between each metric and the handshake parameters of the optimized handshakes. 
As can be expected, a strong positive correlation is observed between the amplitude error and the DTW score, intuitively indicating that large amplitude errors lead to higher DTW scores. We also expect and confirm a strong negative correlation between the frequency error and the PLV score. Additionally, both the DTW and PLV scores are correlated to the mean torque and power, corroborating that higher similarity between the human and robot motions leads to lower energy consumption during the interaction.  Regarding the preferred handshake parameters, a negative correlation is observed between the frequency and stiffness, as well as between the frequency and amplitude. This suggests that users generally prefer higher frequencies when combined with lower stiffness or lower amplitude. However, minimal correlation is found between amplitude and stiffness parameters, implying that user preferences for these two parameters are relatively independent of each other. User satisfaction, however, shows a moderate positive correlation with amplitude error, torque, and power, and a negative correlation with amplitude, implying that users prefer to modulate the stiffness by shaking with larger amplitudes than intended, thereby increasing the error and energy used.

\begin{table}[t]
\caption{Mean and standard deviation of the metrics for the three handshake categories: learning, testing, and optimized. Results are shown for female (F), male (M), and total~(T) participant groups. 
}
\vspace{-0.25em}
\resizebox{1\linewidth}{!}{
\begin{tabular}{ c c c c c}
\hline
Metrics & Gender & Learning Phase & Testing & Optimized \\ \hline
\multirow{3}{*}{\parbox{1.5cm}{\centering Amplitude error [\%]}} & F      & 75.80 ± 62.11    & 46.85 ± 24.51       & 38.62 ± 10.50     \\  
                           & M      & 67.87 ± 60.92    & 64.05 ± 32.56      & 57.81 ± 11.97    \\
                          & T      & 71.36 ± 61.44   & 56.48 ± 29.02       & 49.36 ± 11.33      \\ \hline

\multirow{3}{*}{\parbox{1.5cm}{\centering Frequency error [\%]}}    & F      & 47.22 ± 37.41     & 35.24 ± 20.13       & 15.41 ± 4.45      \\ 
                                    & M      & 34.62 ± 31.94    & 38.74 ± 21.87      & 14.28 ± 4.66      \\ 
                                    & T      & 40.16 ± 34.35     & 37.20 ± 21.10       & 14.78 ± 4.56      \\ \hline
\multirow{3}{*}{DTW [m]}    & F      & 57.69 ± 30.40     & 34.58 ± 10.49       & 32.05 ± 14.79      \\ 
                            & M      & 51.17 ± 28.11     & 46.06 ± 14.09       & 43.77 ± 18.14      \\ 
                            & T      & 54.04 ± 29.12     & 41.01 ± 12.50       & 38.61 ± 5.90      \\ \hline

\multirow{3}{*}{\parbox{1.5cm}{\centering Mean torque [Nm]} }    & F      & 3.00 ± 0.93     & 2.51 ± 0.86       & 2.50 ± 0.74      \\ 
                            & M      & 2.43 ± 0.89     & 2.02 ± 0.87       & 2.17 ± 0.81      \\ 
                            & T      & 2.67 ± 0.92    & 2.23 ± 0.87       & 2.22 ± 0.80      \\ \hline
\end{tabular}}
\label{tab:metrics}
\vspace{-2.1em}
\end{table}

\subsubsection{Note on Passive Handshakes}
We investigate the preferences of active vs.~passive handshakes and the corresponding role of stiffness, similarly to~\cite{mura2020role}. Of the pairwise comparisons that involved one passive and one active handshake, users selected the passive handshake as their preferred option 58\% of the time. Upon further analysis, the average $K_p$ value of the preferred passive handshakes was 86, while the average $K_p$ value of the discarded passive handshakes was 130, suggesting a preference for low stiffness passive handshakes where subjects lead the handshake with the robot
following.

%%%%%%%%%%%%%%%%%%%%%%%%%%%%%%%%%%%%%%%%%%%%%%%%%%%%%%%%%%%%%%%%%%%%%
\section{Conclusion}
\label{sec:conclusion}

In this work, we have presented a method for learning user-preferred parameters for performing handshakes with quadruped robots. We parameterized handshakes with three parameters (amplitude, frequency, and stiffness), and used preference learning to identify user preferences by receiving user feedback following pairwise handshake comparisons. Our results show that this method is effective in identifying optimal user-preferred parameters, with 19/25 participants being happy with their learned handshake parameters, and 5/25 feeling neutral. Moreover, compared with random and test handshakes, the optimized handshakes have significantly decreased human-in-the-loop errors in amplitude and frequency, lower DTW scores, and lower mean torque.

In future work we plan to explore additional methods for user feedback to learn the reward function, including from facial expressions and incorporating power minimization, as an alternative to the verbal binary selection. We will also investigate optimizing additional parameters such as handshake duration or nominal position, which could also play an important role in the overall handshake experience. 

%%%%%%%%%%%%%%%%%%%%%%%%%%%%%%%%%%%%%%%%%%%%%%%%%%%%%%%%%%%%%%%%%%%%%%%%%%%%%%%%
\bibliographystyle{IEEEtran}
\bibliography{refs}

\begin{thebibliography}{10}
\providecommand{\url}[1]{#1}
\csname url@samestyle\endcsname
\providecommand{\newblock}{\relax}
\providecommand{\bibinfo}[2]{#2}
\providecommand{\BIBentrySTDinterwordspacing}{\spaceskip=0pt\relax}
\providecommand{\BIBentryALTinterwordstretchfactor}{4}
\providecommand{\BIBentryALTinterwordspacing}{\spaceskip=\fontdimen2\font plus
\BIBentryALTinterwordstretchfactor\fontdimen3\font minus \fontdimen4\font\relax}
\providecommand{\BIBforeignlanguage}[2]{{%
\expandafter\ifx\csname l@#1\endcsname\relax
\typeout{** WARNING: IEEEtran.bst: No hyphenation pattern has been}%
\typeout{** loaded for the language `#1'. Using the pattern for}%
\typeout{** the default language instead.}%
\else
\language=\csname l@#1\endcsname
\fi
#2}}
\providecommand{\BIBdecl}{\relax}
\BIBdecl

\bibitem{melnyk_physical_2019}
A.~Melnyk and P.~Hénaff, ``\BIBforeignlanguage{en}{Physical {Analysis} of {Handshaking} {Between} {Humans}: {Mutual} {Synchronisation} and {Social} {Context}},'' \emph{\BIBforeignlanguage{en}{International Journal of Social Robotics}}, vol.~11, no.~4, pp. 541--554, Aug. 2019.

\bibitem{chaplin2000handshaking}
W.~F. Chaplin, J.~B. Phillips, J.~D. Brown, N.~R. Clanton, and J.~L. Stein, ``Handshaking, gender, personality, and first impressions.'' \emph{Journal of personality and social psychology}, vol.~79, no.~1, p. 110, 2000.

\bibitem{lafrance1978cultural}
M.~LaFrance and C.~Mayo, ``Cultural aspects of nonverbal communication,'' \emph{International Journal of Intercultural Relations}, vol.~2, no.~1, pp. 71--89, 1978.

\bibitem{katsumi2017nonverbal}
Y.~Katsumi, S.~Kim, K.~Sung, F.~Dolcos, and S.~Dolcos, ``When nonverbal greetings “make it or break it”: the role of ethnicity and gender in the effect of handshake on social appraisals,'' \emph{Journal of Nonverbal Behavior}, vol.~41, pp. 345--365, 2017.

\bibitem{breazeal2004designing}
C.~Breazeal, \emph{Designing sociable robots}.\hskip 1em plus 0.5em minus 0.4em\relax MIT press, 2004.

\bibitem{lee2006physically}
K.~M. Lee, Y.~Jung, J.~Kim, and S.~R. Kim, ``Are physically embodied social agents better than disembodied social agents?: The effects of physical embodiment, tactile interaction, and people's loneliness in human--robot interaction,'' \emph{International journal of human-computer studies}, vol.~64, no.~10, pp. 962--973, 2006.

\bibitem{avelino2018power}
J.~Avelino, F.~Correia, J.~Catarino, P.~Ribeiro, P.~Moreno, A.~Bernardino, and A.~Paiva, ``The power of a hand-shake in human-robot interactions,'' in \emph{2018 IEEE/RSJ international conference on intelligent robots and systems (IROS)}.\hskip 1em plus 0.5em minus 0.4em\relax IEEE, 2018, pp. 1864--1869.

\bibitem{si2016establish}
M.~Si and J.~D. McDaniel, ``Establish trust and express attitude for a non-humanoid robot.'' in \emph{CogSci}, 2016.

\bibitem{otterdijk2023shake}
M.~Van~Otterdijk, D.~Saplacan, A.~Baselizadeh, B.~Laeng, and J.~Torresen, ``To shake or not to shake: Intuitive reactions of senior adults to a robot handshake in a western culture,'' in \emph{2023 32nd IEEE International Conference on Robot and Human Interactive Communication (RO-MAN)}, 2023, pp. 890--896.

\bibitem{prasad2021learning}
V.~Prasad, R.~Stock-Homburg, and J.~Peters, ``Learning human-like hand reaching for human-robot handshaking,'' in \emph{2021 IEEE International Conference on Robotics and Automation (ICRA)}.\hskip 1em plus 0.5em minus 0.4em\relax IEEE, 2021, pp. 3612--3618.

\bibitem{mura2020role}
D.~Mura, E.~Knoop, M.~G. Catalano, G.~Grioli, M.~B{\"a}cher, and A.~Bicchi, ``On the role of stiffness and synchronization in human--robot handshaking,'' \emph{The International Journal of Robotics Research}, vol.~39, no.~14, pp. 1796--1811, 2020.

\bibitem{prasad_human-robot_2022}
V.~Prasad, R.~Stock-Homburg, and J.~Peters, ``\BIBforeignlanguage{en}{Human-{Robot} {Handshaking}: {A} {Review}},'' \emph{\BIBforeignlanguage{en}{International Journal of Social Robotics}}, vol.~14, no.~1, pp. 277--293, Jan. 2022.

\bibitem{knoop2017handshakiness}
E.~Knoop, M.~B{\"a}cher, V.~Wall, R.~Deimel, O.~Brock, and P.~Beardsley, ``Handshakiness: Benchmarking for human-robot hand interactions,'' in \emph{2017 IEEE/RSJ International Conference on Intelligent Robots and Systems (IROS)}.\hskip 1em plus 0.5em minus 0.4em\relax IEEE, 2017, pp. 4982--4989.

\bibitem{artem_physical_2013}
A.~A. Melnyk, M.~V. Khomenko, P.~V. Borysenko, and P.~Hénaff, ``Physical human–robot interaction in the handshaking case: learning of rhythmicity using oscillators neurons,'' \emph{IFAC Proceedings Volumes}, vol.~46, no.~9, pp. 1055--1060, 2013, 7th IFAC Conference on Manufacturing Modelling, Management, and Control.

\bibitem{jouaiti_hebbian_2018}
M.~Jouaiti, L.~Caron, and P.~Hénaff, ``Hebbian {Plasticity} in {CPG} {Controllers} {Facilitates} {Self}-{Synchronization} for {Human}-{Robot} {Handshaking},'' \emph{Frontiers in Neurorobotics}, vol.~12, p.~29, Jun. 2018.

\bibitem{kasuga2005human}
T.~Kasuga and M.~Hashimoto, ``Human-robot handshaking using neural oscillators,'' in \emph{Proceedings of the 2005 IEEE International Conference on Robotics and Automation}.\hskip 1em plus 0.5em minus 0.4em\relax IEEE, 2005, pp. 3802--3807.

\bibitem{beaudoin_haptic_2019}
J.~Beaudoin, T.~Laliberte, and C.~Gosselin, ``\BIBforeignlanguage{en}{Haptic {Interface} for {Handshake} {Emulation}},'' \emph{\BIBforeignlanguage{en}{IEEE Robotics and Automation Letters}}, vol.~4, no.~4, pp. 4124--4130, Oct. 2019.

\bibitem{falahi2014adaptive}
M.~Falahi, T.~A. Shangari, A.~Sheikhjafari, S.~Gharghabi, A.~Ahmadi, and S.~S. Ghidary, ``Adaptive handshaking between humans and robots, using imitation: Based on gender-detection and person recognition,'' in \emph{2014 Second RSI/ISM International Conference on Robotics and Mechatronics (ICRoM)}.\hskip 1em plus 0.5em minus 0.4em\relax IEEE, 2014, pp. 936--941.

\bibitem{christen2019guided}
S.~Christen, S.~Stev{\v{s}}i{\'c}, and O.~Hilliges, ``Guided deep reinforcement learning of control policies for dexterous human-robot interaction,'' in \emph{2019 International Conference on Robotics and Automation (ICRA)}.\hskip 1em plus 0.5em minus 0.4em\relax IEEE, 2019, pp. 2161--2167.

\bibitem{prasad_advances_2020}
V.~Prasad, R.~Stock-Homburg, and J.~Peters, ``Advances in human-robot handshaking,'' in \emph{International Conference on Social Robotics}.\hskip 1em plus 0.5em minus 0.4em\relax Springer, 2020, pp. 478--489.

\bibitem{righetti_dynamic_2006}
L.~Righetti, J.~Buchli, and A.~J. Ijspeert, ``\BIBforeignlanguage{en}{Dynamic {Hebbian} learning in adaptive frequency oscillators},'' \emph{\BIBforeignlanguage{en}{Physica D: Nonlinear Phenomena}}, vol. 216, no.~2, pp. 269--281, Apr. 2006.

\bibitem{miki2022learning}
T.~Miki, J.~Lee, J.~Hwangbo, L.~Wellhausen, V.~Koltun, and M.~Hutter, ``Learning robust perceptive locomotion for quadrupedal robots in the wild,'' \emph{Science Robotics}, 2022.

\bibitem{bellegarda2022cpgrl}
G.~Bellegarda and A.~Ijspeert, ``{CPG-RL}: Learning central pattern generators for quadruped locomotion,'' \emph{IEEE Robotics and Automation Letters}, vol.~7, no.~4, pp. 12\,547--12\,554, 2022.

\bibitem{bellegarda2022robust}
G.~Bellegarda, Y.~Chen, Z.~Liu, and Q.~Nguyen, ``Robust high-speed running for quadruped robots via deep reinforcement learning,'' in \emph{2022 IEEE/RSJ International Conference on Intelligent Robots and Systems (IROS)}, 2022, pp. 10\,364--10\,370.

\bibitem{bellegarda2020robust}
G.~{Bellegarda} and Q.~{Nguyen}, ``Robust quadruped jumping via deep reinforcement learning,'' \emph{arXiv preprint arXiv:2011.07089}, 2020.

\bibitem{shafiee2023deeptransition}
M.~Shafiee, G.~Bellegarda, and A.~Ijspeert, ``Viability leads to the emergence of gait transitions in learning agile quadrupedal locomotion on challenging terrains,'' \emph{Nature Communications}, vol.~15, no.~1, p. 3073, 2024.

\bibitem{shafiee2023puppeteer}
M.~Shafiee, G.~Bellegarda, and A.~Ijspeert, ``Puppeteer and marionette: Learning anticipatory quadrupedal locomotion based on interactions of a central pattern generator and supraspinal drive,'' in \emph{2023 IEEE International Conference on Robotics and Automation (ICRA)}, 2023, pp. 1112--1119.

\bibitem{cheng2023parkour}
X.~Cheng, K.~Shi, A.~Agarwal, and D.~Pathak, ``Extreme parkour with legged robots,'' \emph{arXiv preprint arXiv:2309.14341}, 2023.

\bibitem{unitreeGO1}
{Unitree Robotics}. Go1. \url{https://www.unitree.com/products/go1/}.

\bibitem{tagne_physical_2016}
G.~Tagne, P.~Hénaff, and N.~Gregori, ``Measurement and analysis of physical parameters of the handshake between two persons according to simple social contexts,'' in \emph{2016 IEEE/RSJ International Conference on Intelligent Robots and Systems (IROS)}, 2016, pp. 674--679.

\bibitem{biyik_active_2020}
E.~Biyik, N.~Huynh, M.~Kochenderfer, and D.~Sadigh, ``Active preference-based gaussian process regression for reward learning,'' in \emph{Robotics: Science and Systems}, 2020.

\bibitem{akrour2012april}
R.~Akrour, M.~Schoenauer, and M.~Sebag, ``April: Active preference learning-based reinforcement learning,'' in \emph{Machine Learning and Knowledge Discovery in Databases: European Conference, ECML PKDD 2012.}\hskip 1em plus 0.5em minus 0.4em\relax Springer, 2012, pp. 116--131.

\bibitem{sadigh_active_2017}
D.~Sadigh, A.~Dragan, S.~Sastry, and S.~Seshia, ``Active {Preference}-{Based} {Learning} of {Reward} {Functions},'' in \emph{Robotics: {Science} and {Systems} {XIII}}.\hskip 1em plus 0.5em minus 0.4em\relax Robotics: Science and Systems Foundation, Jul. 2017.

\bibitem{basu2017you}
C.~Basu, Q.~Yang, D.~Hungerman, M.~Singhal, and A.~D. Dragan, ``Do you want your autonomous car to drive like you?'' in \emph{Proceedings of the 2017 ACM/IEEE International Conference on Human-Robot Interaction}, 2017, pp. 417--425.

\bibitem{tucker2020preference}
M.~Tucker, E.~Novoseller, C.~Kann, Y.~Sui, Y.~Yue, J.~W. Burdick, and A.~D. Ames, ``Preference-based learning for exoskeleton gait optimization,'' in \emph{2020 IEEE international conference on robotics and automation (ICRA)}.\hskip 1em plus 0.5em minus 0.4em\relax IEEE, 2020, pp. 2351--2357.

\bibitem{ingraham2022role}
K.~A. Ingraham, C.~D. Remy, and E.~J. Rouse, ``The role of user preference in the customized control of robotic exoskeletons,'' \emph{Science robotics}, vol.~7, no.~64, p. eabj3487, 2022.

\bibitem{lee2023user}
U.~H. Lee, V.~S. Shetty, P.~W. Franks, J.~Tan, G.~Evangelopoulos, S.~Ha, and E.~J. Rouse, ``User preference optimization for control of ankle exoskeletons using sample efficient active learning,'' \emph{Science Robotics}, vol.~8, no.~83, p. eadg3705, 2023.

\bibitem{righetti08}
L.~Righetti and A.~J. Ijspeert, ``Pattern generators with sensory feedback for the control of quadruped locomotion,'' in \emph{IEEE International Conference on Robotics and Automation}, 2008, pp. 819--824.

\bibitem{bellegarda2024visual}
G.~Bellegarda, M.~Shafiee, and A.~Ijspeert, ``Visual {CPG-RL}: Learning central pattern generators for visually-guided quadruped locomotion,'' \emph{arXiv preprint arXiv:2212.14400}, 2024.

\bibitem{bellegarda2024quadruped}
G.~Bellegarda, M.~Shafiee, M.~E. {\"O}zberk, and A.~Ijspeert, ``Quadruped-frog: Rapid online optimization of continuous quadruped jumping,'' \emph{arXiv preprint arXiv:2403.06954}, 2024.

\bibitem{biyik_aprel_2022}
E.~B{\i}y{\i}k, A.~Talati, and D.~Sadigh, ``Aprel: A library for active preference-based reward learning algorithms,'' in \emph{2022 17th ACM/IEEE International Conference on Human-Robot Interaction (HRI)}.\hskip 1em plus 0.5em minus 0.4em\relax IEEE, 2022, pp. 613--617.

\bibitem{berndt1994using}
D.~J. Berndt and J.~Clifford, ``Using dynamic time warping to find patterns in time series,'' in \emph{Proceedings of the 3rd international conference on knowledge discovery and data mining}, 1994, pp. 359--370.

\bibitem{orefice_lets_2016}
P.-H. Orefice, M.~Ammi, M.~Hafez, and A.~Tapus, ``Let's handshake and i'll know who you are: Gender and personality discrimination in human-human and human-robot handshaking interaction,'' in \emph{2016 IEEE-RAS 16th International Conference on Humanoid Robots (Humanoids)}, 2016, pp. 958--965.

\end{thebibliography}

\end{document}